\documentclass{article}

     \usepackage[preprint,nonatbib]{neurips_2019}

\usepackage[utf8]{inputenc} 
\usepackage[T1]{fontenc}    
\usepackage{hyperref}       
\usepackage{url}            
\usepackage{booktabs}       
\usepackage{amsfonts}
\usepackage{amsmath}
\usepackage{amssymb}
\usepackage{nicefrac}       
\usepackage{microtype}      
\usepackage{authblk}
\usepackage{graphicx}
\usepackage{subcaption}
\usepackage{caption}
\usepackage{placeins}

\title{$\rho$-VAE: \\Autoregressive parametrization of the VAE encoder}

\author{\textbf{Sohrab Ferdowsi}}
\author{\textbf{Maurits Diephuis}}
\author{\textbf{Shideh Rezaeifar}}
\author{\textbf{Slava Voloshynovkiy}}

\affil{
Department of Computer Science, University of Geneva, Switzerland \authorcr
  \{\tt sohrab.ferdowsi, maurits.diephuis, shideh.rezaeifar, svolos\}@unige.ch}

\begin{document}

\maketitle

\begin{abstract}

We make a minimal, but very effective alteration to the VAE model. This is about a drop-in replacement for the (sample-dependent) approximate posterior to change it from the standard white Gaussian with diagonal covariance to the first-order autoregressive Gaussian. We argue that this is a more reasonable choice to adopt for natural signals like images, as it does not force the existing correlation in the data to disappear in the posterior. Moreover, it allows more freedom for the approximate posterior to match the true posterior. This allows for the repararametrization trick, as well as the KL-divergence term to still have closed-form expressions, obviating the need for its sample-based estimation. Although providing more freedom to adapt to correlated distributions, our parametrization has even less number of parameters than the diagonal covariance, as it requires only two scalars, $\rho$ and $s$, to characterize correlation and scaling, respectively. As validated by the experiments, our proposition noticeably and consistently improves the quality of image generation in a plug-and-play manner, needing no further parameter tuning, and across all setups. The code to reproduce our experiments is available at \url{https://github.com/sssohrab/rho_VAE/}.
 
\end{abstract}

\section{Introduction}
Arguably, one of the most successful approaches to generative modeling and representation learning is that of ``Auto-encoding variational Bayes'' \cite{VAE}. Considering a latent-based model for the data, where some underlying but hidden variations are assumed to be responsible for the creation of the observed data, this approach realizes the standard variational Bayes in the form of a neural network and offers a practical recipe for end-to-end learning of its parameters, while providing effective approximation of the intractable posterior. This has then given rise to the very popular Variational AutoEncoder (VAE) framework, a family of models successful at generating high quality images (e.g., see \cite{dai2019diagnosing}, \cite{rezende14} , \cite{LossyVAE} and \cite{PixelVAE} among others), as well as learning useful representation with little or no supervision (e.g., as in \cite{kingma2014semi}).

Essentially, the VAE bridges the tasks of generation of the data from latent codes, with that of inferring the latent codes from the data as two parts of the same body: the decoder and the encoder of an autoencoder architecture, respectively. This should then induce an implicit statistical model for each of these parts.

As for the decoder, the standard model is a white Gaussian distribution, centered on the latent codes when passed through the decoder. To provide higher capacity and hence matching better with natural images, this can then be generalized to autoregressive models which bring about better performance, e.g., as in \cite{FlowVAE, gulrajani2016pixelvae, lucas2018auxiliary}, albeit adding to the computational burden.

The encoder part, however, is more delicate to treat. From one hand, it has to be realistic enough to match the true posterior and hence tighten the variational bound. From the other hand, it should be kept simple to make the gradient-based optimization feasible. This has multiple requirements: Firstly, it is preferred to have a closed-form expression for the regularization of the posterior to push it closer to the prior. Even by chosing a simple prior, this is usually not possible.\footnote{Not to mention that, in order to produce high quality images, the prior as well may be needed to be more complicated, e.g., as in \cite{VampPrior}.} Secondly, since the link between the encoder and the decoder cannot be direct, as the explicit generation of latent codes breaks the differentiability, the encoder should be parametrized such that it can be injected to the latent space through the reparametrization trick \cite{VAE}. 

The standard solution, however, favors more the side of pragmatism by choosing the easy diagonal Gaussian distribution as the encoder's approximate posterior. Being insufficient in practice, a wealth of efforts\footnote{The reader is encouraged to consult good reviews like \cite{tschannen2018recent} that provide overviews of the VAE literature.} have been targeted ever since to address some of its issues like failing to perform amortized inference or not learning very meaningful latent features. As examples of some of these efforts, $\beta$-VAE \cite{higgins2017beta} alters the optimization towards more regularization in order to encourage the latent space to be more factorized in the hope that it would result in disentanglement. Another attempt is the info-VAE \cite{InfoVAE} which addresses the mismatch between the powerful decoder and the non-flexible encoder by involving more terms to the optimization.

But before making any such complementary attempts and focusing again on the basic VAE model, this work proposes a very easy solution that can effectively improve the encoding quality of any VAE model, where relevant. Our solution is as pragmatic as the standard VAE encoding and runs as fast, but is much more flexible. Moreover, it is equally compatible with all further attempts to improve the VAE's, e.g., $\beta$-VAE or info-VAE, as we will experimentally corroborate, thanks to its realization as a drop-in replacement for the standard approximate posterior. 

Next in in section \ref{sec:VAE}, we briefly review the standard VAE model highlighting aspects relevant to our work. Our proposition, the $\rho$-VAE is then introduced in section \ref{sec:proposed}, on which we perform experiments in section \ref{sec:exp}. The paper is finally concluded in section \ref{sec:conclusions}.

\section{The standard VAE model} \label{sec:VAE}
In a typical probabilistic model where a latent variable $\mathbf{z} \in \Re^d$ is the underlying factor to generate the observable samples $\mathbf{x}$'s $\in \Re^n$, the standard variational Bayes \cite{jordan1999introduction} paradigm is concerned with finding an approximation $q(\mathbf{z})$ for the intractable posterior $p(\mathbf{z}|\mathbf{x})$. This is achieved by minimizing the Kullback-Leibler divergence between these two distributions , i.e., $D_{\text{KL}}\big[ q(z) || p(z|x) \big]$. Standard treatments of this quantity, along with its non-negativity property will then amount to the following inequality:

\begin{equation}  \label{eq:VB_ELBO}
\log(p(\mathbf{x})) \leqslant \mathbb{E}_{q(\mathbf{z})} \Big[ \log(p(\mathbf{x}|\mathbf{z}))  \Big] - D_{\text{KL}}\Big[ q(\mathbf{z}) || p(\mathbf{z}) \Big].
\end{equation}

Autoencoding variational Bayes \cite{VAE} is then constructing an explicit dependence of the latent variables to the $i^{\text{th}}$ training sample by considering a parametrized distribution $q_{\phi}(\mathbf{z}^{(i)}|\mathbf{x}^{(i)})$ for the approximate posterior, whose construction resembles the encoder part of an autoencoder network with a set of learnable weights $\phi$. Furthermore, the training samples can be decoded with $p_{\theta}(\mathbf{x}^{(i)} | \mathbf{z}^{(i)})$, another network with parameters symbolized as $\theta$.

Making this double-sided data dependence more explicit, and by summing over all $N$ training samples results to the following inequality:
\begin{equation} \label{eq:vae_ELBO}
\frac{1}{N}\sum_{i=1}^N \log(p(\mathbf{x}^{(i)})) \leqslant \frac{1}{N} \sum_{i=1}^N \Big[ \log(p_{\theta}(\mathbf{x}^{(i)}|\mathbf{z}^{(i)}))   -  D_{\text{KL}}\big[ q_{\phi}(\mathbf{z}^{(i)}  | \mathbf{x}^{(i)}) || p(\mathbf{z}) \big] \Big].
\end{equation}

This, in fact, is highly relevant for generative modeling as the marginal log-likelihood of the training samples will be upper bounded by two terms, both of which amenable to mini-batch optimization with stochastic gradient descent. 

During optimization, the first term of the LHS can be considered as a data fidelity term, minimized e.g., in the $\ell_2$ sense, since a natural choice for the decoder is $p_{\theta}(\mathbf{x}^{(i)}|\mathbf{z}^{(i)}) = \mathcal{N} \Big( \big( \mathbf{x}^{(i)}|\mathbf{z}^{(i)} \big) , \sigma^2 \mathrm{I}_n \Big)$, where $\mathrm{I}_n$ is the $n$-dimensional unity matrix.

The second term, from the other hand, can be interpreted as a regularization term, pushing the approximate posterior to a prior imposed on the latent space, most conveniently a simple $p(z) = \mathcal{N}(\mathbf{0}, \mathrm{I}_d)$. Provided that the optimization is successful, and the inequality (\ref{eq:vae_ELBO}) is tight, one can generate random samples from this prior, pass it through the learned decoder and generate samples (non-trivially) similar to the underlying data.   

However, the above scenario comes with a major caution: the fact that sampling $\mathbf{z}$ from $q_{\phi}(\mathbf{z} | \mathbf{x})$ is a non-differentiable operation. The work-around for this issue is the wise ``reparametrization trick'', as proposed in \cite{VAE}. 

The idea is to create the required randomness from a fixed distribution $\boldsymbol{\epsilon} \sim \mathcal{N}(\mathbf{0}, \mathrm{I}_d)$. The samples of the appropriate distribution can then be generated by injecting the learnable moments, e.g., using $\mathbf{z}^{(i)} = \boldsymbol{\mu}^{(i)} + \tilde{\mathrm{C}}^{(i)} \boldsymbol{\epsilon}$, where $\boldsymbol{\mu}^{(i)}$ is the mean vector of the posterior learned for the $i^{\text{th}}$ sample and $\tilde{\mathrm{C}}^{(i)}$ is the Choleskiy decomposition of the corresponding covariance matrix $\mathrm{C}^{(i)}$.

This then limits the practical choices for $\mathrm{C}^{(i)}$ to have analytical Choleskiy decomposition forms, since both $\mathrm{C}^{(i)}$ and $\tilde{\mathrm{C}}^{(i)}$ participate in the optimization simultaneously.

Another issue to address is the calculation of $D_{\text{KL}}\big[ q_{\phi}(\mathbf{z}^{(i)}  | \mathbf{x}^{(i)}) || p(\mathbf{z}) \big]$. While there are several choices (e.g., replacing the KL-divergence with other variants, or the adversarial density ratio trick \cite{nguyen2010estimating}), in order to avoid many practical difficulties, the standard choice is to pick a closed-form expression for it, hence further limiting the choices of $\mathrm{C}^{(i)}$.

While the prior distribution is chosen as $p(z) = \mathcal{N}(\mathbf{0}, \mathrm{I}_d)$, considering the above two constraints, the standard choice widely adopted in many further variants for the sample-wise approximate posterior is to set $\mathrm{C}_{(\mathbf{s})}^{(i)} = \text{diag} \big( \mathbf{s}^{(i)} \big)$. In other words,  $q_{\phi}(\mathbf{z}^{(i)}  | \mathbf{x}^{(i)}) = \mathcal{N} \big( \boldsymbol{\mu}^{(i)}, \text{diag} \big( \mathbf{s}^{(i)} \big)  \big)$, a diagonal Gaussian distribution parametrized by the pair$(\boldsymbol{\mu}^{(i)}, \mathbf{s}^{(i)})$.

Note that now, the reparametrization trick can run smoothly, since the Choleskiy decomposition has a closed expression as $\tilde{\mathrm{C}}_{(\mathbf{s})}^{(i)} = \text{diag} \big( \sqrt{\mathbf{s}^{(i)}} \big) $. Furthermore, the regularization term $D_{\text{KL}}\big[ q_{\phi}(\mathbf{z}^{(i)}  | \mathbf{x}^{(i)}) || p(\mathbf{z}) \big] \Big]$ is also calculated analytically as:
\begin{equation}  \label{eq:vae_KLD_loss}
D_{\text{KL}}\Big[ \mathcal{N} \Big( \boldsymbol{\mu}^{(i)}, \text{diag} \big( \mathbf{s}^{(i)} \big) \Big)    \Big|\Big|  \mathcal{N} \big( \mathbf{0}, \mathrm{I}_d \big)  \Big] = \frac{1}{2} \Big[ \mathbf{1}_d^T \mathbf{s}^{(i)} + \big|\big| \boldsymbol{\mu}^{(i)} \big|\big|_2^2 -d -  \mathbf{1}_d^T \log \big({\mathbf{s}^{(i)}} \big) \Big], 
\end{equation}
where $||\cdot||_2^2$ is the squared $\ell_2$-norm, and $\log \big({\mathbf{s}^{(i)}} \big)$ is applied element-wise.

While this is a very practical choice, we argue in section \ref{sec:proposed} that it is too restrictive, as it disregards any correlation within dimensions.
 
\section{The $\rho$-VAE}  \label{sec:proposed}
We saw that two considerations limit the choices of approximate posterior: the need for a parametric Choleskiy factorization of its covariance matrix, as well as closed-form expression for the regularization term of (\ref{eq:vae_ELBO}), which basically requires the expression of log-determinant  of the covariance.

In spite of the general consensus to pick $\mathrm{C}_{(\mathbf{s})}^{(i)} = \text{diag}\big( \mathbf{s}^{(i)} \big)$, which does not allow any correlation between the dimensions of the approximate posterior, this work proposes another parametrization that grants such freedom, satisfies the above-mentioned restrictions, and yet has less number of parameters.

In particular, we chose a first-order autoregressive covariance which is characterized by a scaling factor $s$, and another scalar $\rho$ to control the level of correlation, hence the term $\rho$-VAE. This has the form of a simple symmetric Toeplitz matrix as the following:

\begin{equation}  \label{eq:rho_cov}
\mathrm{C}_{(\rho,s)} = s \times  \text{Toeplitz} \Big([1,\rho, \rho^2, \cdots, \rho^{d-1}] \Big)
= s \begin{bmatrix}
    1          & \rho        & \rho^2     & \rho^3       & \cdots   & \rho^{d-1} \\
    \rho       & 1           & \rho       & \rho^2       & \cdots   & \rho^{d-2} \\
    \rho^2     & \rho        & 1          & \rho         & \cdots   & \rho^{d-3} \\
    \rho^3     & \rho^2      & \rho       & 1            & \cdots   & \rho^{d-4} \\
    \vdots     &       &       & \ddots       & \ddots   & \vdots      \\
    \rho^{d-1} & \cdots  & \rho^3 & \rho^2   & \rho   & 1
  \end{bmatrix},
\end{equation}

where $s$ is a positive scalar, and the correlation parameter is bounded as $-1 < \rho  < +1$.

The determinant for this matrix can be calculated as \cite{10.2307/1993228}: 

\begin{equation} \label{eq:rho_det}
\text{det} \big( \mathrm{C}_{(\rho,s)} \big) = s^d (1 - \rho^2)^{d-1},
\end{equation}
based on which we can derive the regularization term of the loss function as:

\begin{equation} \label{eq:rho_KLD_loss}
D_{\text{KL}}\Big[ \mathcal{N} \Big( \boldsymbol{\mu}^{(i)}, \mathrm{C}_{(\rho,s)} \Big)    \Big|\Big|  \mathcal{N} \big( \mathbf{0}, \mathrm{I}_d \big)  \Big] = \frac{1}{2} \Big[
\big|\big| \boldsymbol{\mu}^{(i)} \big|\big|_2^2 + d(s-1-\log{(s)})  - (d-1)\log{(1 - \rho^2)}  
\Big].
\end{equation}

As far as the reparametrization trick is concerned, the Choleskiy decomposition of our choice of covariance matrix has the following lower triangular form:

\begin{equation}    \label{eq:rho_chol}
\tilde{\mathrm{C}}_{(\rho,s)} = \frac{1}{\sqrt{s}} \begin{bmatrix}
    1          & 0        & 0     & 0       & 0   & 0 \\
    \rho       & \sqrt{1-\rho 2}           & 0       & 0       & \cdots   & 0 \\
    \rho^2     & \rho\sqrt{1-\rho 2}        & \sqrt{1-\rho 2}          & 0         & \cdots   & 0 \\
    \rho^3     & \rho^2\sqrt{1-\rho 2}      & \rho\sqrt{1-\rho 2}       & \sqrt{1-\rho 2}           & \cdots   & 0 \\
    \vdots     &       &       & \ddots       & \ddots   & \vdots      \\
    \rho^{d} & \cdots  & \rho^3 \sqrt{1-\rho 2} & \rho^2 \sqrt{1-\rho 2}   & \rho \sqrt{1-\rho 2}   & \sqrt{1-\rho 2}
  \end{bmatrix},
\end{equation}
which can be used to generate the latent codes as $\mathbf{z}^{(i)} = \boldsymbol{\mu}^{(i)} + \tilde{\mathrm{C}}_{(\rho,s)}^{(i)} \boldsymbol{\epsilon}$, which can be constructed also as the element-wise product of $\mathrm{C}_{(\rho,s)}$ with another highly structured matrix.

Otherwise, if depending on the choice of the deep learning framework used, the realization of Toeplitz matrices is not straightforward, one can generate AR(1) samples directly from their definition, i.e., $\mathbf{z}^{(i)}[j] = \boldsymbol{\mu}^{(i)}[j] + \sqrt{s} \boldsymbol{\epsilon}[j] + \rho \mathbf{z}^{(i)}[j-1]$, for $1 < j \leqslant d$.

Although it has less number of parameters than the standard choice and is hence more resilient towards over-fitting, this structure for the approximate posterior is more natural to consider, since correlation will somehow be represented. 

Note that the fact that the prior is chosen as a white Gaussian by design, i.e., $p(\mathbf{z}) = \mathcal{N}(\mathbf{0}, \mathrm{I}_d)$, does not obviate the need for the per-sample approximate posterior to account for correlation. In fact, the per-sample posterior can be correlated, yet the aggregation of all samples can be a white Gaussian matching the prior.

More importantly, the need for correlation does not solely stem from the natural signals like images being correlated. As a matter of fact, another requirement for the success of the VAE-based generative modeling is the tightness of the bound in (\ref{eq:vae_ELBO}), which is controlled by $D_{\text{KL}}\big[ q_{\phi}(\mathbf{z}^{(i)}  | \mathbf{x}^{(i)}) || p_{\theta}(\mathbf{z}^{(i)} | \mathbf{x}^{(i)}) \big]$.

In other words, to guarantee a successful training, the approximate posterior should have enough capacity to match the unknown and intractable posterior. In VAE models, however, it is usually only ``hoped'' that this will be the case. We believe (albeit without providing quantitative evidence), that accounting for correlation may help reduce this gap.

Next we will show the effectiveness of our proposition. We show that the simple alterations to the standard approach, without the need for any sort of hyper-parameter tuning will noticeably and consistently improve the performance under all variations considered and for all setups. 

\section{Experiments} \label{sec:exp} 

We perform experiments on 4 variants of the VAE and across the mnist \cite{lecun-mnisthandwrittendigit-2010} and the fashion \cite{xiao2017fashion} databases. For each of these models, we first use the diagonal Gaussian approximate posterior as the baseline and then replace it with our AR(1) proposition. As was explained in section \ref{sec:proposed}, this only changes the reparametrization step, as well as the closed-form expression of the regularizer.

In order to force the network to output correlation factors in the range $-1 < \rho < 1$, we pass the output of the corresponding linear layer (mapping from $\Re^{d'}$ to $\Re$, where $d'$ is the dimension of an intermediate hidden-layer) through the $\tanh(\cdot)$ activation. To ensure positive scaling of the covariance, similar to the standard implementations, we consider the output of the corresponding linear layer (from $\Re^{d'}$ to $\Re$ in our case) as $\log(s)$, and exponentiate it where necessary.\footnote{So instead of the standard linear layer of size $d' \times d$, we have two $d' \times 1$ linear layers.}

Other than these adjustments, we keep every other thing\footnote{For example the dimensions, regularization constants, network architectures, learning rates, epochs, ...} exactly the same. 

\begin{figure}  [!h]
   \begin{center} 
\includegraphics[width=1.0\textwidth]{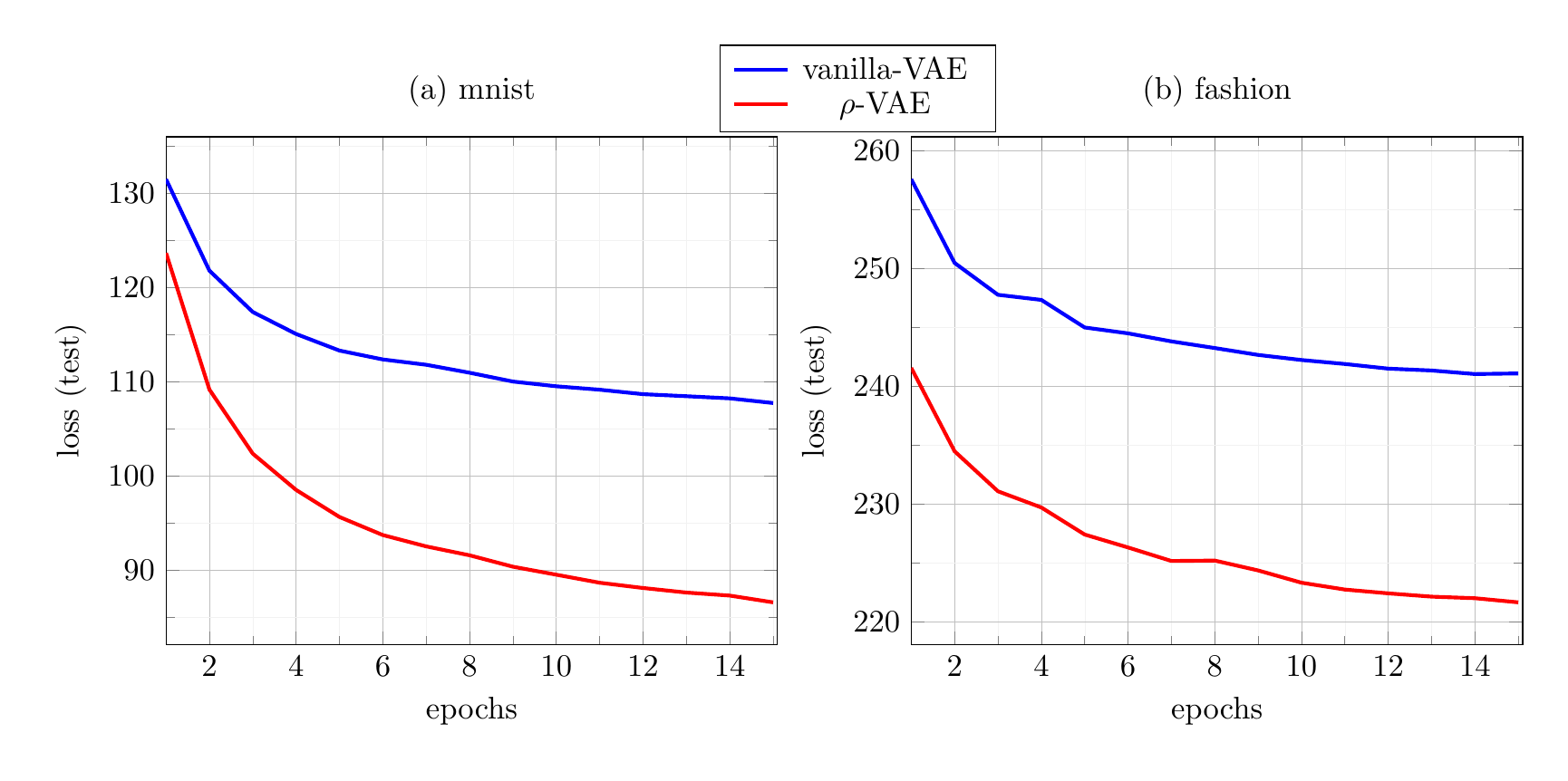}

\end{center}
\vspace{-.75cm}    
   \caption{The loss function profile of the test set for the vanilla-VAE and $\rho$-VAE models on (a) mnist and (b) fashion databases.}
   \label{fig:curve_VanillaVAE}
   \end{figure}

\begin{figure}  [!h]
   \begin{center} 
\subcaptionbox{vanilla-VAE (mnist)} {\includegraphics[width=0.47\textwidth]{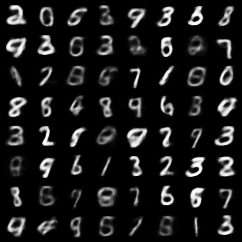}} 
\subcaptionbox{vanilla-VAE (fashion)} {\includegraphics[width=0.47\textwidth]{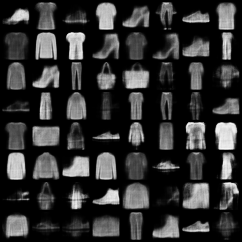}} 

\subcaptionbox{$\rho$-VAE (mnist)} {\includegraphics[width=0.47\textwidth]{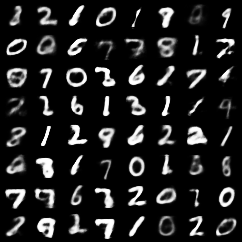}} 
\subcaptionbox{$\rho$-VAE (fashion)} {\includegraphics[width=0.47\textwidth]{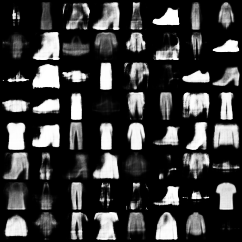}} 

\end{center}
\vspace{-0.4cm}    
   \caption{Randomly generated samples from the vanilla-VAE and the $\rho$-VAE models (no cherry-picking of the samples).}
   \label{fig:pic_VanillaVAE}
   \end{figure}

Figure \ref{fig:curve_VanillaVAE} shows the loss function, i.e., the lower-bound on the negative log-likelihood (-LHS of (\ref{eq:vae_ELBO})) for the vanilla-, as well as the $\rho$-VAE, where a clear advantage is seen for the latter on both databases.\footnote{Note that our figures show the loss function on the test set starting from the end of the first epoch, where already several mini-batches of optimization have been run. So both variants start from the same loss before starting the optimization and they are measuring the same quantity.} This is then confirmed visually in Fig. \ref{fig:pic_VanillaVAE}, where the samples generated from $\rho$-VAE look much sharper than the baseline. 

 \begin{figure}  [!h]
   \begin{center} 
\includegraphics[width=1.0\textwidth]{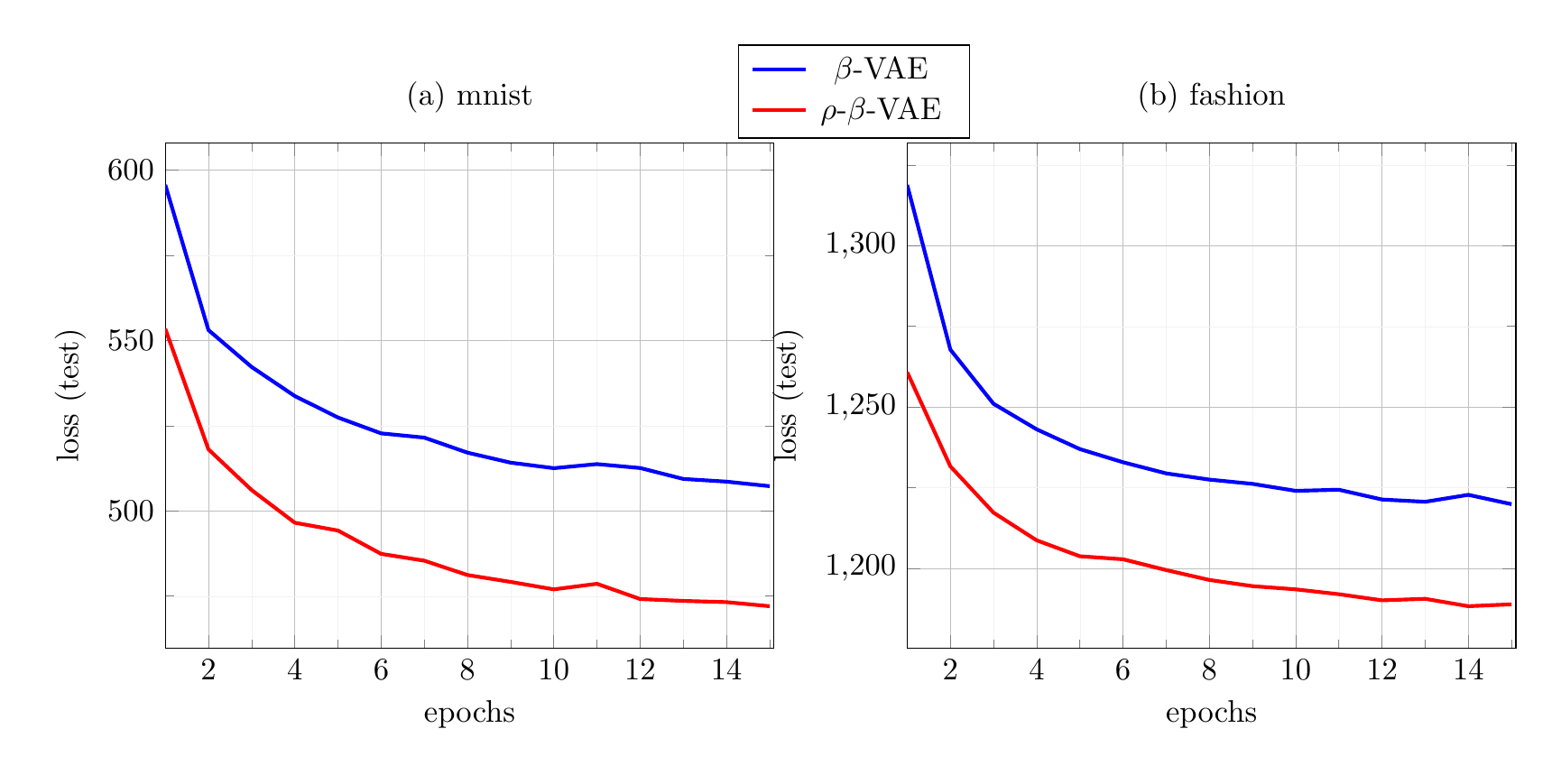}
\end{center}
\vspace{-.75cm}     
   \caption{The loss function profile of the test set for the $\beta$-VAE and $\rho$-$\beta$-VAE models on (a) mnist and (b) fashion databases.}
   \label{fig:curve_BETAVAE}
   \end{figure}

Figures \ref{fig:curve_BETAVAE} shows the loss function profilefor the two variants on the $\beta$-VAE framework \cite{higgins2017beta}, respectively. Note that, as argued by the authors, in order to encourage factorization of the latent, the $\beta$-VAE scales the loss function in favor of the regularization term. Here, too, we show a consistent gap of performance in favor of the$\rho-$ variant.\footnote{In fact, we observe in our experiments (not shown here) that the gain in performance is due both to a decreased reconstruction loss, as well as lower KL-divergence, and hence keeping the advantage similar in the $\beta$-VAE case.}

\begin{figure}  [!t]
   \begin{center} 
\includegraphics[width=1.0\textwidth]{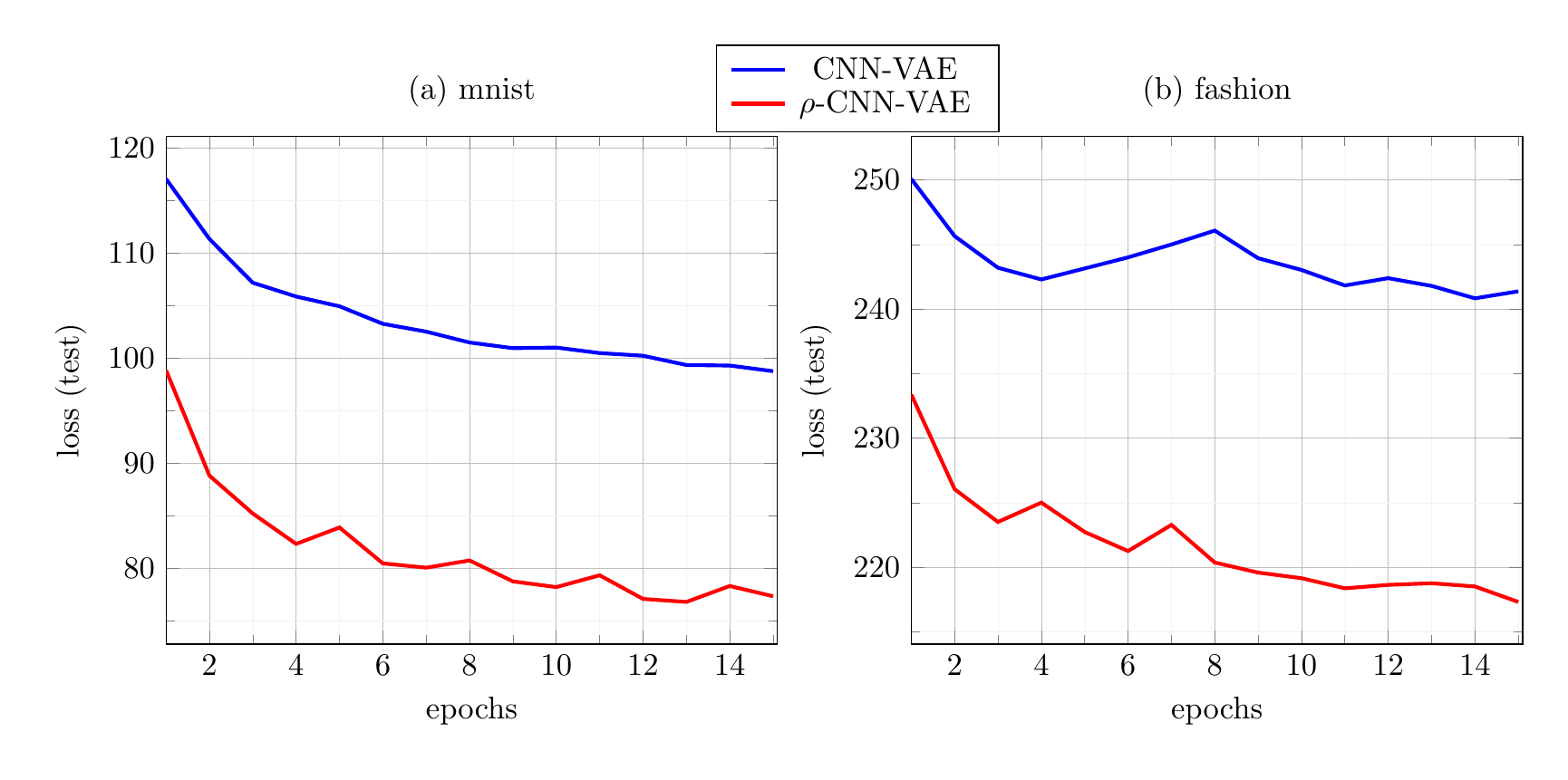}
\end{center}
\vspace{-.75cm}   
   \caption{The loss function profile of the test set for the CNN-VAE and $\rho$-CNN-VAE models on (a) mnist and (b) fashion databases.}
   \label{fig:curve_CNNVAE}
   \end{figure}

We next perform our comparison on a VAE model with convolutions. This has two convolutional layers with $64$ and $128$ filters of size $4 \times 4 $, and then two fully-connected layers to produce the hidden intermediate layer. From the latent code, then the decoding is done in a symmetric way with transposed convolutions. Again we see clear advantage by the adoption of the AR(1) structure in the network in figure \ref{fig:curve_CNNVAE}.

%

 \begin{figure}  [!t]
   \begin{center} 
\includegraphics[width=1.0\textwidth]{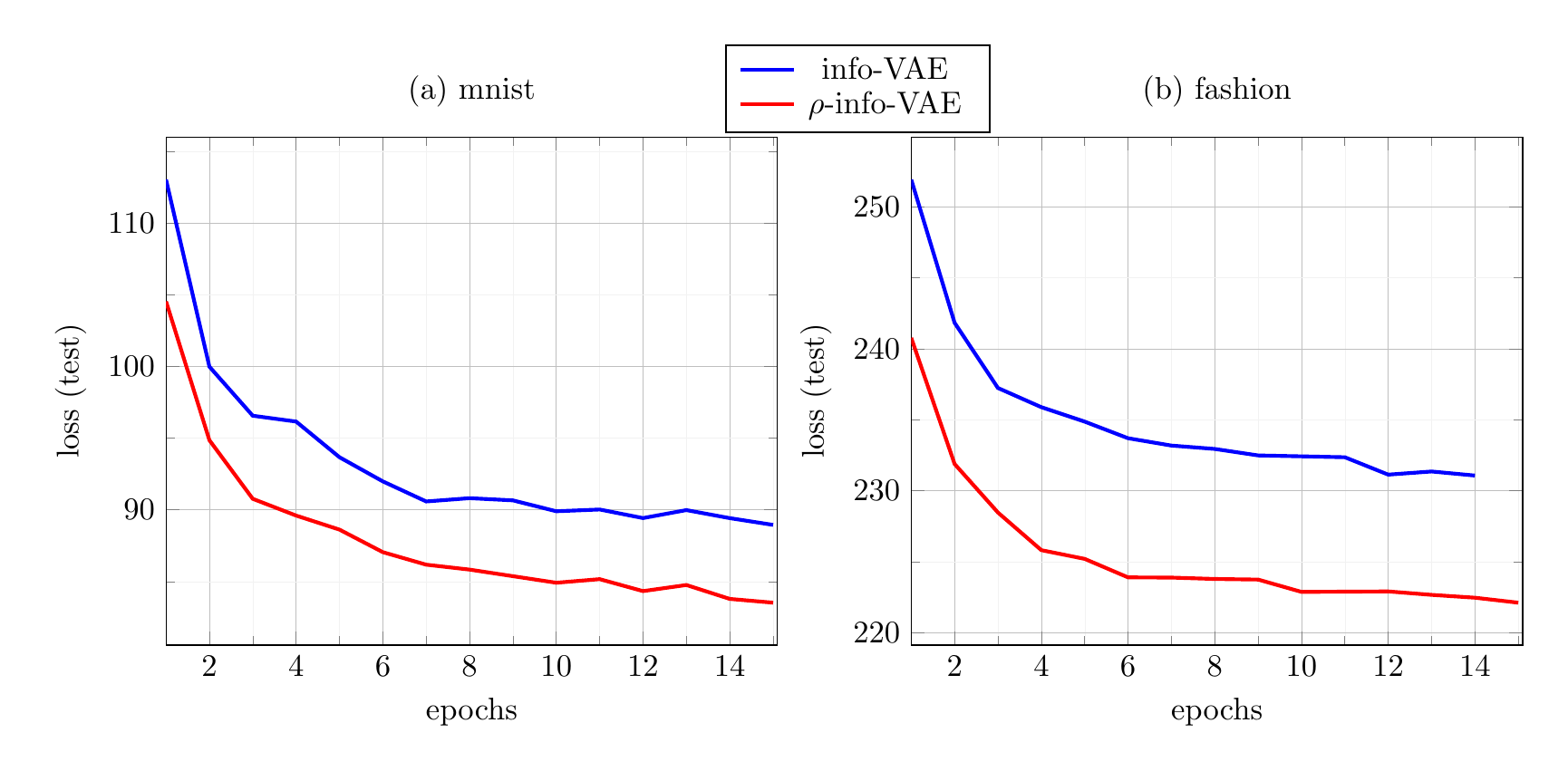}
\end{center}
\vspace{-.75cm}  
   \caption{The loss function profile of the test set for the info-VAE and $\rho$-info-VAE models on (a) mnist and (b) fashion databases. Both models use DC-GAN encoding and decoding}
   \label{fig:curve_INFOVAE}
   \end{figure}

As for the last experiment, we take our $\rho$-parametrization and plug it into the info-VAE model \cite{InfoVAE}. This has a more complicated optimization that further involves the aggregated approximate posterior, for which we use the DC-GAN \cite{radford2015unsupervised}, as proposed by the authors for the two variants. Again we observe more successful training in Figure \ref{fig:curve_INFOVAE}.

\section{Conclusions} \label{sec:conclusions}
We proposed to replace the standard and ubiquitous parametrization of the approximate posterior within VAE models as diagonal Gaussian with the  much more flexible AR(1) Gaussian distribution. We argued that this choice, not only does not add any complexity or issue to the optimization, but it even has less number of parameters and can be easily integrated in other VAE models in a plug-and-play manner. Being able to let correlation to propagate within the approximate posterior, we argued that it might match better to the true posterior. Our proposition showed consistent improvement to the quality of image generation, both quantitatively and visually with sharper and samples. 
\FloatBarrier

\bibliographystyle{unsrt}

\bibliography{Bibliography}

\end{document}